\documentclass[10pt,letterpaper]{article}

\usepackage{cogsci}
\usepackage{graphicx}
\usepackage{booktabs}
\cogscifinalcopy 
\newcommand{\YZ}[1]{\textcolor{black}{#1}}

\usepackage[
  style=apa,
  natbib=true,
  annotation=false,
]{biblatex}
\usepackage{float} 

\title{Modeling Sarcastic Speech: Semantic and Prosodic Cues in a Speech Synthesis Framework}

\author[1]{\mbox{Zhu Li}}
\author[2]{\mbox{Yuqing Zhang}}
\author[1]{\mbox{Xiyuan Gao}}
\author[1]{\mbox{Shekhar Nayak}}
\author[1]{\mbox{Matt Coler}}
\affil[1]{Speech Technology Lab, University of Groningen, Campus Fryslân, The Netherlands}
\affil[2]{Center for Language and Cognition, University of Groningen, The Netherlands}

\begin{document}

\maketitle

\begin{abstract}
Sarcasm is a pragmatic phenomenon in which speakers convey meanings that diverge from literal content, relying on an interaction between semantics and prosodic expression. However, how these cues jointly contribute to the recognition of sarcasm remains poorly understood.
We propose a computational framework that models sarcasm as the integration of semantic interpretation and prosodic realization. Semantic cues are derived from an LLaMA 3 model fine-tuned to capture discourse-level markers of sarcastic intent, while prosodic cues are extracted through semantically aligned utterances drawn from a database of sarcastic speech, providing prosodic exemplars of sarcastic delivery.
Using a speech synthesis testbed, perceptual evaluations show that semantic and prosodic cues enhance perceived sarcasm, with the combined system achieving the best downstream F1 while maintaining high subjective sarcasm ratings.
These findings highlight the complementary roles of semantics and prosody in pragmatic interpretation and illustrate how modeling can shed light on the mechanisms underlying sarcastic communication.

\textbf{Keywords:}
pragmatics; sarcasm; prosody; semantic cues; speech perception
\end{abstract}

\section{Introduction}
Sarcasm plays an important role in everyday communication by enabling speakers to express mockery, criticism, or contempt through meanings that diverge from literal interpretation. As a pragmatic phenomenon, sarcasm is often realized through irony, where ostensibly positive or neutral expressions convey an underlying negative intent. While such ironic intent can sometimes be inferred from explicit linguistic content, sarcasm frequently lacks clear lexical markers and therefore cannot be reliably identified from semantics alone. Instead, its interpretation depends on a nuanced interplay of semantic incongruity, contextual information, and prosodic cues. Speakers commonly employ prosodic strategies, such as exaggerated intonation, emphatic stress, or lengthened syllables, to signal sarcastic intent, particularly in cases where linguistic cues are subtle or ambiguous \citep{rockwell2000lower, cheang2008sound, li2024functional}. This reliance on multiple cues makes sarcasm a 
complex phenomenon that challenges 
computational modeling, as successful automatic recognition requires integrating information across modalities.
Understanding how listeners integrate these cues is a fundamental question for theories of pragmatic language comprehension.

One challenge in studying sarcastic communication lies in systematically manipulating semantic and prosodic cues while holding other factors constant. Traditional corpus-based or behavioral studies often lack fine-grained control over semantic and prosodic cues, since these dimensions tend to be entangled in natural speech and are difficult to manipulate independently or in a graded manner as experimental stimuli \citep{bryant2002recognizing, bryant2010prosodic, rockwell2007vocal}. Speech synthesis, however, offers a powerful experimental testbed, enabling controlled generation of utterances in which semantic and prosodic information can be independently varied to explore their respective contributions to sarcasm perception.

Recent advances in computational modeling provide new opportunities to formally represent and manipulate the cues underlying pragmatic communication.
Large language models (LLMs) have shown an ability to capture discourse-level semantic and pragmatic patterns, including forms of semantic subtlety relevant to sarcasm. At the same time, retrieval-based approaches allow models to condition generation on prosodic exemplars that reflect expressive sarcasm patterns. Together, these methods offer a way to formalize hypotheses about how semantic content and prosodic realization jointly contribute to sarcastic communication.

In this study, we introduce a sarcasm-aware speech synthesis framework that models sarcasm as the integration of semantic content and prosodic expression. 
We independently manipulate two types of cues to condition sarcastic speech generation in our synthesis framework: 1) semantic representations of sarcastic intent derived from a fine-tuned LLM, and 2) prosodic patterns drawn from aligned reference utterances in a curated speech database of sarcastic speech. Using perceptual evaluations of the synthesized speech, we assess how each cue contributes to listeners' perception of sarcasm.
Perceptual evaluations reveal that both cues independently enhance sarcasm recognition, with the strongest effects emerging when they are combined. These findings provide computational evidence for the complementary roles of semantics and prosody in pragmatic interpretation and demonstrate how synthesis-based models can serve as tools for modeling speech behavior.

\section{Related Work}
\subsection{Sarcasm Detection and Modeling}
In recent years, substantial progress has been made in computational sarcasm detection. Many studies have adopted multimodal approaches that integrate textual, acoustic, and visual cues, demonstrating significant improvements over unimodal methods \citep{castro-etal-2019-towards, ray-etal-2022-multimodal, raghuvanshi2025intra, gao2024amused, li2025evaluating}. Datasets such as MUStARD \citep{castro-etal-2019-towards} and its extensions \citep{ray-etal-2022-multimodal} have facilitated this line of work by providing aligned textual, auditory, and visual data for sarcasm recognition. While these efforts highlight the feasibility of modeling sarcasm computationally, they have largely focused on classification and recognition tasks. 
\YZ{Recent work has quantified how pragmatic information is distributed across text and prosody. 
For example, \citet{yadavalli2025prosody} show that prosody carries substantial information about sarcasm when long-term discourse context is unavailable.}
However, relatively little attention has been paid to the generation of sarcastic speech or to using computational models as tools for exploring how different cues contribute to sarcasm perception. Moving from detection to synthesis is not only a natural next step but also a crucial one, as it enables interactive systems to actively deploy pragmatic phenomena rather than passively identify them.

\subsection{Expressive Speech Synthesis and Sarcasm}
Research on expressive text-to-speech (TTS) synthesis has primarily concentrated on broad emotional categories such as happiness, anger, and sadness \citep{wang2018style, akuzawa2018expressive, li2021controllable}. Although these systems can produce emotionally expressive and intelligible speech, more subtle 
and pragmatic phenomena remain largely unexplored. Sarcasm, despite being widely used in everyday communication, has received particularly little attention in speech synthesis research.
However, sarcastic or other pragmatic speech synthesis holds considerable potential for improving human-computer interaction in applications such as conversational agents and entertainment systems \citep{ritschel2019irony}. 
Prior work has attempted to analyze acoustic correlates of sarcastic speech, identifying prosodic features related to pitch, pace, and loudness \citep{cheang2008sound, li2024functional}. However, only a limited number of studies have explored directly manipulating such prosodic features to generate sarcastic-sounding speech \citep{peters2017creating}. 
This difficulty arises in part because sarcasm is inherently more subtle and context-dependent than conventional emotions such as anger or joy, making it difficult to capture with handcrafted acoustic features or simple prosodic controls. Moreover, the development of sarcastic speech synthesis systems is hindered by the scarcity of high-quality, annotated sarcastic speech corpora, which limits the applicability of purely data-driven approaches \citep{li2023sarcasticspeech, li2025integrating}.

\subsection{LLM-Based Semantics and Retrieval-Based Speech Synthesis}
Recent advances in natural language processing have introduced new opportunities for modeling pragmatic phenomena. LLMs have demonstrated the ability to capture high-level semantic and discourse-level information. In the TTS domain, integrating LLM-derived embeddings has been shown to enable more nuanced prosody control and improve emotional expressiveness and contextual appropriateness \citep{feng2024llama, li_pl-tts_2024, shen2025get}. Such semantic representations are particularly relevant for sarcasm, which often depends on pragmatic incongruity between literal meaning and intended attitude. Leveraging LLM-based semantic representations could therefore provide a principled way of conditioning TTS models on sarcasm-relevant cues that handcrafted features fail to capture.

Complementary to semantic modeling, retrieval-based methods have been explored as a way to condition speech generation on external knowledge or examples retrieved from large databases. Retrieval-augmented generation (RAG) has been widely applied in text-based tasks such as question answering and dialogue \citep{lewis2020retrieval, shuster2021retrieval}, and related ideas have been adopted in speech synthesis through reference-based and style-transfer approaches, where reference utterances are used to guide prosodic realization \citep{xue_retrieval_2024, luo_autostyle-tts_2025}. However, existing approaches often rely on manually selected or speaker-specific reference audio, limiting scalability and contextual alignment. In the context of sarcasm, where annotated corpora are scarce, automatic retrieval of semantically similar sarcastic speech could provide prosodic exemplars that both enrich training and guide generation in data-scarce settings. This suggests a paradigm in which LLMs supply semantic-pragmatic cues while the RAG module retrieves expressive references for prosodic encoding, together enabling a systematic modeling of subtle pragmatic phenomena such as sarcasm.

\begin{figure*}[htbp]
    \centering
    \includegraphics[width=0.95\linewidth]{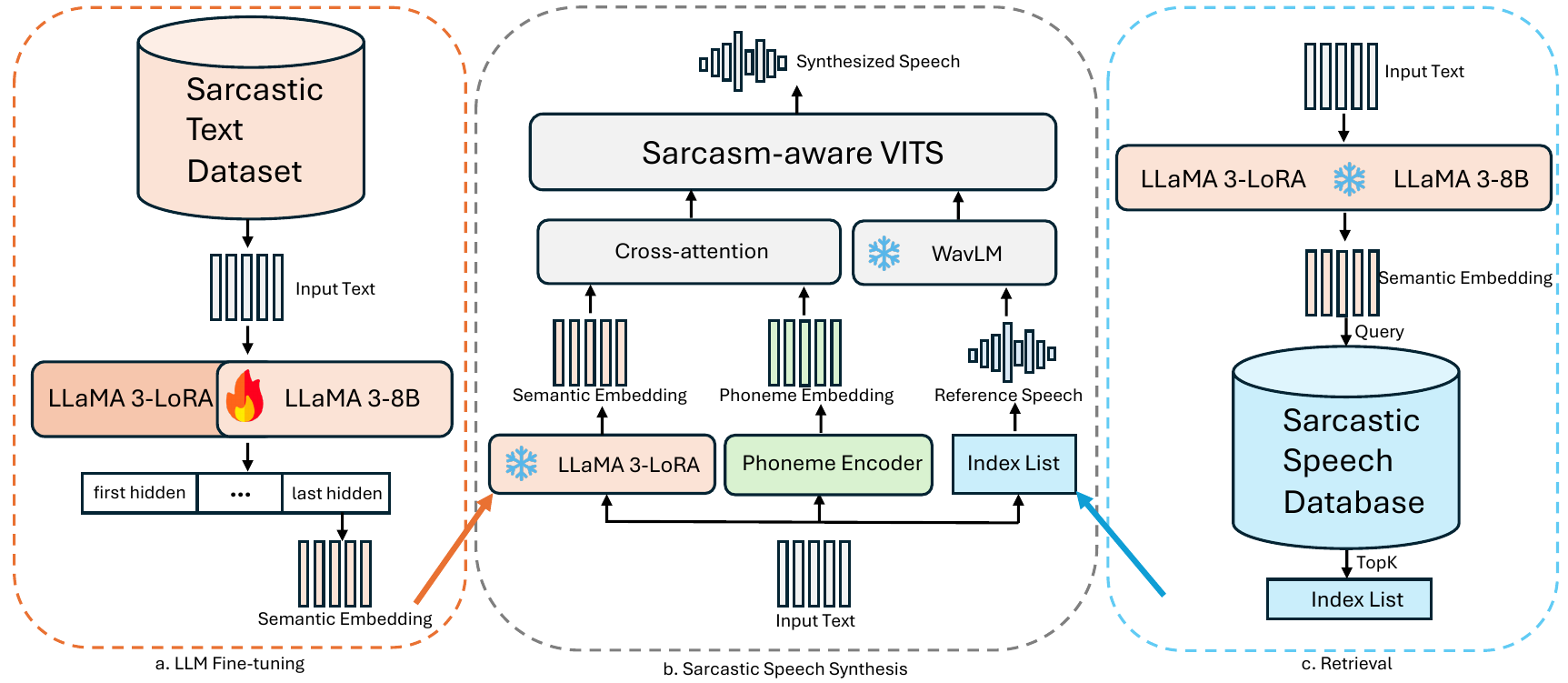}
    \caption{Overview of the proposed framework modeling sarcasm as an interaction between semantic and prosodic cues. 
Semantic representations capturing semantic and pragmatic cues are extracted from a sarcasm-adapted LLaMA 3 model, while prosodic exemplars are retrieved from reference sarcastic speech. 
These cues are integrated within a speech synthesis architecture to generate utterances that convey sarcastic intent.}
    \label{fig:RAGLLMTTS}
\end{figure*}

\section{Methods}

We propose a Retrieval-Augmented LLM-enhanced sarcastic speech modeling framework. An overview of the architecture is illustrated in Figure~\ref{fig:RAGLLMTTS}.

\subsection{Sarcasm-Aware Semantic Encoding via Parameter-Efficient Fine-tuning (PEFT)}
\label{sec:lora}
To systematically manipulate semantic cues of sarcasm, we encode input text into sarcasm-aware semantic representations using a PEFT method on a sarcasm-labeled corpus. This allows us to isolate and represent high-level semantic features indicative of sarcastic intent for subsequent synthesis and perceptual evaluation.

Low-Rank Adaptation (LoRA) has recently been widely adopted for adapting LLMs across diverse domains such as emotion recognition, sentiment analysis, and affective detection, due to its parameter efficiency and effectiveness \citep{cai2024lora}. 
Motivated by these advances, we leverage LoRA to finetune the LLaMA 3-8B model for sarcasm-aware semantic encoding. Fine-tuning is performed on a curated sarcasm-labeled dataset, enabling the model to capture signals of sarcastic intent, including pragmatic and discourse-level cues.

Formally, given an input text sequence $x = (w_1, \dots, w_{T_t})$ of length $T_t$, the LLaMA 3-LoRA encoder produces a sequence of contextualized semantic embeddings:
\begin{equation}
\mathbf{E}_s = f_{\text{LoRA}}(x) \in \mathbb{R}^{T_t \times d_t},
\end{equation}
where $d_t$ denotes the hidden dimensionality of the language model. Each row of $\mathbf{E}_s$ corresponds to a token-level representation extracted from the hidden state of the final transformer layer, encoding meanings relevant to sarcasm and providing a controllable semantic representation for sarcasm synthesis.

\subsection{RAG for Prosody Conditioning}
To investigate the role of prosodic cues in sarcasm perception, we retrieve semantically aligned sarcastic utterances as prosodic exemplars. These exemplars allow us to manipulate prosody independently of semantic content for synthesis.

We integrate an RAG module into the sarcastic speech synthesis framework. 
The key idea is to leverage semantically aligned sarcastic utterances as prosodic references, providing the generated speech with realistic intonation patterns.
Concretely, we first construct an index $\mathcal{D} = \{(u_i, \mathbf{a}_i)\}_{i=1}^N$ of sarcastic utterances $u_i$ curated from MUStARD++, where $\mathbf{a}_i$ denotes their pre-computed semantic representations using the same LLaMA 3-LoRA adapted LLaMA 3-8B encoder. 
For a given input text, the semantic embedding $\mathbf{E}_s$ produced by the LLaMA 3-LoRA encoder is used as a retrieval query to fetch the top-$K$ semantically relevant sarcastic utterances from the indexed database; in this work, we set $K=3$.

We first compute the cosine similarity between $\mathbf{E}_s$ and each database entry:
\begin{equation}
\text{sim}(\mathbf{E}_s, \mathbf{a}_i) 
= \frac{\mathbf{E}_s \cdot \mathbf{a}_i}{\|\mathbf{E}_s\|\|\mathbf{a}_i\|}.
\end{equation}

We then retrieve the top-$K$ most relevant utterances. Each retrieved utterance $u_k \in \mathcal{U}_{\text{top-}K}$ is encoded by WavLM \citep{chen2022wavlm} to obtain a prosody embedding: 
\begin{equation}
\mathbf{E}_{w_k} = \text{Pool}\!\left(f_{\text{WavLM}}(u_k)\right) \in \mathbb{R}^{d_w},
\end{equation}
where $\text{Pool}(\cdot)$ aggregates the frame-level features into a fixed-length vector, and $d_w$ denotes the dimensionality of the resulting prosodic embedding. These exemplars provide style references that reflect the characteristic intonation and emphasis patterns of sarcastic speech. 

Compared to conventional style transfer methods that rely on a single manually chosen reference utterance, the proposed retrieval mechanism offers two key advantages. First, it scales more naturally to diverse input contexts, since the retrieval process automatically identifies semantically aligned exemplars. 
Second, by conditioning on multiple retrieved samples, the model can capture a richer variety of sarcastic prosodic patterns, leading to more contextually appropriate speech.

\subsection{LLM-Enhanced and Retrieval-Guided TTS for Sarcasm}

The system builds upon the VITS architecture \citep{kim2021conditional}, enriched with semantic features extracted by a fine-tuned LLaMA 3 and guided by prosodic exemplars retrieved via a RAG module. 

Let $\mathbf{E}_p \in \mathbb{R}^{T_p \times d_p}$ denote the sequence of phoneme embeddings, where $T_p$ is the number of phonemes in the input sequence and $d_p$ is the dimensionality
of the phoneme embedding space. Let $\mathbf{E}_s \in \mathbb{R}^{T_t \times d_t}$ denote the sarcasm-aware semantic textual embeddings produced by the fine-tuned LLaMA~3.
In the cross-attention module, we treat $\mathbf{E}_p$ as the query and $\mathbf{E}_s$ as key and value:

\begin{equation}
\mathbf{Q} = \mathbf{E}_p \mathbf{W}_q, \quad
\mathbf{K} = \mathbf{E}_s \mathbf{W}_k, \quad
\mathbf{V} = \mathbf{E}_s \mathbf{W}_v,
\end{equation}

\begin{equation}
\mathbf{H} = \text{Softmax}\Big(\frac{\mathbf{Q}\mathbf{K}^\top}{\sqrt{d_k}}\Big)\mathbf{V},
\end{equation}

where $\mathbf{W}_q, \mathbf{W}_k, \mathbf{W}_v$ are learned projection matrices, and $d_k$ is the dimensionality of the key. The resulting cross-attention output $\mathbf{H}$ represents a phoneme-aligned semantic encoding.

Finally, the prosody exemplars $\mathcal{E}_w = \{\mathbf{E}_{w_1}, \dots, \mathbf{E}_{w_K}\}$ extracted by WavLM are linearly projected and added to the cross-attention output over semantic and phoneme embeddings to modulate the decoder:

\begin{equation}
\mathbf{Z} = \mathbf{H} + \sum_{k=1}^K \mathbf{W}_w \mathbf{E}_{w_k},
\end{equation}

where $\mathbf{W}_w$ is a learned linear projection mapping the prosody embeddings to the decoder dimension. This yields hidden states $\mathbf{Z}$ that are conditioned on phonemes, semantic content, and retrieved prosodic cues, enabling the synthesis of sarcastic speech with expressive intonation.

\section{Experiments}

To investigate how semantic and prosodic cues contribute to the perception of sarcasm, we conducted a series of experiments using synthesized speech with systematically manipulated cues. We define a set of experimental conditions that independently and jointly vary semantic and prosodic information. Specifically, we examine: 
\begin{itemize}
    \item \textbf{Baseline}: speech generated without sarcasm-specific semantic or prosodic guidance;
    \item \textbf{Semantic-only}: speech conditioned on sarcasm-aware semantic embeddings derived from a fine-tuned LLaMA~3 model;
    \item \textbf{Prosody-only}: speech guided by prosodic exemplars retrieved from a database of sarcastic utterances;
    \item \textbf{Combined}: speech incorporating both semantic and prosodic guidance.
\end{itemize}

In later analyses, we further explore sub-variants within the semantic-only condition, including embeddings from BERT, pretrained LLaMA~3, and LoRA-fine-tuned LLaMA~3, in order to assess the effect of different semantic representations. These conditions collectively allow us to evaluate the relative contributions of semantic cues, prosodic expression, and their interaction to listeners’ perception of sarcastic intent. 

\subsection{Data}

For the pre-training phase of VITS, we use the HiFi-TTS corpus \citep{bakhturina2021hi}, a high-fidelity audiobook dataset with diverse speakers and wideband recordings.

To build the sarcastic speech database for retrieval purposes, we use the MUStARD++ dataset \citep{ray-etal-2022-multimodal}. MUStARD++ is a multimodal sarcasm detection corpus compiled from popular sitcoms such as Friends and The Big Bang Theory. It contains 1,202 audiovisual utterances, equally divided into 601 sarcastic and 601 non-sarcastic samples. 
Following standard practice, we partition the corpus into training, validation, and test sets with an 8:1:1 ratio.
Specifically, 10\% of the data is held out for evaluation, while the remaining 80\% of sarcastic samples are used to construct the retrieval-based sarcastic speech database. This setup ensures relatively sufficient coverage of sarcastic prosody in the retrieval dataset.

To adapt LLaMA 3 as sarcasm-aware semantic encoders via LoRA, we fine-tune them on the News Headlines Sarcasm dataset \citep{misra2023Sarcasm}. This dataset contains over 28,000 headlines from sources such as \textit{The Onion}, each annotated with a binary sarcasm label. The class distribution is roughly balanced, making it suitable for learning discourse-level markers of sarcastic intent. The LLM serves as a feature extractor that provides high-level semantic embeddings conditioned on sarcastic intent. 
To validate the effectiveness of this adaptation, we train a sarcasm detection model using various textual embeddings, including BERT, LLaMA 3, LLaMA 3-LoRA, and compare their performance as evidence of each embedding's ability to capture sarcastic intent.

\subsection{Experimental Setup}

For training the baseline VITS model, we follow the standard configuration described in \citet{kim2021conditional}. The model is pre-trained on the HiFi-TTS corpus \citep{bakhturina2021hi} using the open-source Amphion toolkit\footnote{\url{https://github.com/open-mmlab/Amphion}}.

To adapt LLMs for sarcasm-aware synthesis, the LoRA fine-tuning of LLaMA 3 is performed with an expansion factor of 8 and a learning rate of $1e-4$, updating only the low-rank adapters in the attention layers while freezing the backbone weights\footnote{\url{https://github.com/hiyouga/LLaMA-Factory}}.

\subsection{Compared Methods}

To systematically assess the contribution of semantic and prosodic cues to sarcasm perception, we synthesized speech under six conditions corresponding to different manipulations of these cues:

\begin{itemize}
    \item \textbf{Baseline}: Speech generated using the standard variational text-to-speech model VITS, without any sarcasm-specific semantic or prosodic guidance, serving as a neutral reference.
    \item \textbf{Semantic-only (BERT)}: Speech conditioned on semantic embeddings from a pretrained BERT model \citep{devlin2019bert}, providing general contextual semantic features. 
    \item \textbf{Semantic-only (LLaMA 3)}: Speech conditioned on embeddings from LLaMA 3 \citep{dubey2024llama3}, enabling richer contextual and pragmatic information. 
    \item \textbf{Semantic-only (LLaMA 3-LoRA)}: Speech conditioned on sarcasm-aware semantic embeddings from a LoRA-fine-tuned LLaMA 3 model, capturing discourse-level cues indicative of sarcastic intent.
    \item \textbf{Prosody-only (RAG)}: Speech guided by prosodic exemplars retrieved from a database of sarcastic utterances, providing realistic intonation patterns while keeping semantic content neutral.
    \item \textbf{Semantic + Prosody (Proposed)}: Speech generated using both sarcasm-aware semantic embeddings (LLaMA 3-LoRA) and retrieved prosodic exemplars (RAG), allowing joint manipulation of semantic incongruity and prosodic expression to maximize perceived sarcasm.
\end{itemize}

\section{Results and Discussion}

\subsection{Effect of Different Semantic Embeddings on Sarcasm Intent Modeling}

We first evaluate 
how well different semantic embeddings capture semantic cues associated with sarcastic intent.
Specifically, we train and evaluate sarcasm detection models on the text transcripts of the MUStARD++ dataset, using the same 8:1:1 data split and following the same experimental setup as \citet{ray-etal-2022-multimodal}. The goal of this experiment is to assess how effectively different semantic embeddings capture sarcasm-relevant cues and to motivate the choice of embeddings used later for conditioning speech synthesis. Models are evaluated with Precision (P), Recall (R), and weighted F1-score (F1), with results summarized in Table~\ref{tab:sarcasm_detection}.

\begin{table}[h!]
\centering 
\caption{Performance of different models on sarcasm detection. 
MUStARD++ (text) refers to the original sarcasm detection systems used in \citet{ray-etal-2022-multimodal}.}
\label{tab:sarcasm_detection}
\begin{tabular}{lccc}
\toprule
\textbf{Method} & \textbf{P (\%)} & \textbf{R (\%)} & \textbf{F1 (\%)} \\
\midrule
MUStARD++ (text) & 67.9 & 67.7 & 67.7 \\
BERT             & 66.8 & 66.9 & 66.8 \\
LLaMA 3          & 65.7 & 65.4 & 65.5 \\
LLaMA 3-LoRA     & \textbf{72.4} & \textbf{72.7} & \textbf{72.5} \\
\bottomrule
\end{tabular}
\end{table}

Both the MUStARD++ baseline and BERT achieved moderate performance with F1-scores around 67\%. LLaMA 3 performed slightly worse (65.5\%), suggesting that general-purpose pretraining alone is insufficient to capture the nuanced cues of sarcasm. In contrast, PEFT with LoRA yielded a clear improvement, raising the F1-score to 72.5\%. This indicates that targeted fine-tuning effectively operationalizes sarcasm-relevant semantic features.
These results provide evidence that semantic embeddings from the LoRA-enhanced model more reliably encode discourse-level markers of sarcasm. This motivates their use in the TTS framework: by conditioning speech synthesis on embeddings that better reflect sarcastic intent, we can systematically examine whether improvements in semantic cue representation translate to perceptually stronger expressions of sarcasm in synthesized speech.

\subsection{Sarcastic Speech Synthesis Results}

\begin{table*}[t]
\centering
\caption{Evaluation of sarcastic speech synthesis systems. 
Objective acoustic metrics (MCD, pitch, energy) assess signal fidelity; 
sarcasm detection (Precision (P), Recall (R), and weighted F1-score (F1)) evaluates how well sarcastic intent can be recovered from synthesized speech; 
subjective ratings measure perceived naturalness (NMOS) and sarcasm expressivity (SMOS).}
\label{tab:tts_results}
\resizebox{\textwidth}{!}{
\begin{tabular}{lcc ccc ccc cc}
\toprule
& \multicolumn{2}{c}{\textbf{Cue Type}} 
& \multicolumn{3}{c}{\textbf{Objective}} 
& \multicolumn{3}{c}{\textbf{Sarcasm Detection}} 
& \multicolumn{2}{c}{\textbf{Subjective}} \\
\cmidrule(lr){2-3}
\cmidrule(lr){4-6}
\cmidrule(lr){7-9}
\cmidrule(lr){10-11}
\textbf{Method} 
& \textbf{Semantic} 
& \textbf{Prosodic} 
& \textbf{MCD $\downarrow$} 
& \textbf{Pitch $\downarrow$} 
& \textbf{Energy $\downarrow$} 
& \textbf{P $\uparrow$} 
& \textbf{R $\uparrow$} 
& \textbf{F1 $\uparrow$} 
& \textbf{NMOS $\uparrow$} 
& \textbf{SMOS $\uparrow$} \\
\midrule
Ground Truth 
& -- & -- 
& -- & -- & -- 
& 62.8 & 62.3 & 62.3 
& $3.8 \pm 0.1$ 
& $4.5 \pm 0.2$ \\
\midrule
Baseline 
& -- & -- 
& $9.8$ & $261.9$ & $4.4$
& 60.3 & 60.5 & 59.9 
& $2.6 \pm 0.1$ 
& $3.2 \pm 0.2$ \\

w/ BERT
& \checkmark & -- 
& $\textbf{9.6}$ & $265.6$ & $4.5$ 
& 60.5 & 60.6 & 60.5 
& $2.5 \pm 0.1$ 
& $3.1 \pm 0.2$ \\

w/ LLaMA 3 
& \checkmark & -- 
& $10.4$ & $261.6$ & $4.4$ 
& 59.6 & 59.7 & 59.0 
& $2.0 \pm 0.1$ 
& $2.6 \pm 0.2$ \\

w/ LLaMA 3-LoRA
& \checkmark & -- 
& $10.1$ & $282.6$ & $4.4$ 
& 60.6 & 60.8 & 60.6 
& $2.6 \pm 0.1$ 
& $3.8 \pm 0.2$ \\

w/ RAG
& -- & \checkmark 
& $10.0$ & $261.0$ & $4.4$ 
& 61.5 & 61.7 & 61.6 
& $2.6 \pm 0.1$ 
& $3.7 \pm 0.2$ \\

w/ LoRA + RAG
& \checkmark & \checkmark 
& $9.8$ & $\textbf{259.6}$ & $4.4$ 
& \textbf{62.7} & \textbf{62.9} & \textbf{62.5} 
& $\mathbf{2.7 \pm 0.1}$ 
& $\mathbf{3.8 \pm 0.2}$ \\
\bottomrule
\end{tabular}}
\end{table*}

Table~\ref{tab:tts_results} summarizes the evaluation of synthesized speech under different cue manipulations: semantic embeddings, prosodic exemplars, and their combination. We consider three perspectives: 1) low-level acoustic consistency, measured by mel-cepstral distortion (MCD), pitch root mean square error (RMSE), energy RMSE; 2) operationalized sarcasm expressivity, evaluated via a downstream sarcasm detection model applied to synthesized audio, and 3) subjective perception of sarcasm, assessed through two mean opinion score (MOS) tests: naturalness MOS (NMOS) and sarcasm MOS (SMOS). To assess downstream sarcasm detection performance, we adopt the detection architecture used in MUStARD++ with collaborative gating \citep{ray-etal-2022-multimodal}, using only the \textit{speech modality}. Features extracted from synthesized audio are compared against target sarcasm labels, where higher agreement indicates stronger sarcasm expressiveness.


Overall, speech generated with both sarcasm-aware semantic embeddings (LoRA-fine-tuned LLaMA 3) and retrieved prosodic exemplars exhibited the strongest performance across measures. Specifically, the combined condition achieved a comparably lower distortion rate (9.83 MCD) and the most stable prosodic statistics, while also yielding the highest downstream sarcasm detection F1-score (62.5\%). 
Semantic embeddings from the untuned LLaMA 3 alone resulted in higher distortion and lower detection performance, suggesting that general-purpose embeddings fail to fully capture the semantic cues critical for conveying sarcasm. Fine-tuning with LoRA improves semantic alignment, and further incorporating prosodic exemplars enhances the perceptual expressivity of the synthesized utterances.
Importantly, low-level acoustic metrics (MCD, pitch, energy) remained relatively stable across conditions, indicating that improvements in sarcasm expressivity are primarily due to high-level semantic and prosodic cue manipulation, rather than changes in basic signal fidelity. 
Moreover, the increases observed in sarcasm detection closely mirror the subjective ratings, suggesting that the objective recognizability of sarcastic intent reliably translates to perceptual expressivity, as evaluated below.

To complement the measures above, we conducted a perceptual study to evaluate how well the synthesized speech conveyed sarcasm and naturalness. We recruited 30 listeners from diverse backgrounds\footnote{We did not collect demographic variables beyond English proficiency and hearing status; we acknowledge this as a limitation.} who were asked to rate randomly shuffled samples from each system on two dimensions: (i) perceived naturalness (NMOS) and (ii) perceived sarcasm expressivity (SMOS), both on 5-point Likert scales.
Each participant listened to eight sentences generated by six systems, plus the ground truth, resulting in a total of 56 utterances\footnote{Inter-rater reliability for naturalness ratings was modest at the individual-rater level (ICC$_{(2,30)} = .86$ for NMOS and (ICC$_{(2,30)} = .80$ for SMOS).}. Pairwise comparison across conditions reveals several important trends. Baseline speech without semantic or prosodic guidance was generally intelligible but relatively flat, receiving moderate naturalness (2.6 ± 0.1) and sarcasm expressivity (3.2 ± 0.2). Adding general-purpose semantic embeddings (BERT) did not meaningfully improve ratings (2.5 ± 0.1 NMOS, 3.1 ± 0.2 SMOS), suggesting that embeddings lacking sarcasm-specific information may be insufficient to strengthen perceptual sarcasm cues. 
In addition, directly integrating LLaMA 3 degraded performance ($2.0 \pm 0.1$ NMOS, $2.6 \pm 0.2$ SMOS). Listeners frequently reported that these samples sounded less natural and had flat intonation patterns. This indicates that raw LLM embeddings are poorly aligned with intended expressive speech. 
In contrast, semantic embeddings adapted for sarcastic content were associated with higher perceived sarcasm ratings (3.8 ± 0.2 SMOS) while maintaining naturalness comparable to the baseline (2.6 ± 0.1 NMOS). Listeners consistently reported clearer prosodic patterns, such as exaggerated intonation, that aligned with the intended sarcastic meaning, suggesting that targeted semantic conditioning via LoRA adaptation can operationalize semantically and pragmatically relevant sarcastic cues in speech.
Incorporating prosodic exemplars retrieved from real sarcastic utterances also yielded relatively high sarcasm expressivity ratings (SMOS 3.7–3.8), without reducing naturalness ($2.6 \pm 0.1$ NMOS). Samples were described as having intonation and emphasis patterns more consistent with authentic sarcastic speech, making the intended sarcasm more immediately recognizable by a large portion of participants.
Overall, the combined manipulation of semantic and prosodic cues achieved the highest downstream sarcasm-detection F1 and the highest numerical NMOS ($2.7 \pm 0.1$ NMOS) among synthesized systems, while maintaining high perceived sarcasm expressivity ($3.8 \pm 0.2$ SMOS). However, because its SMOS is comparable to the LoRA-only condition, we interpret the subjective results as evidence for complementary cue contributions rather than a clear additive perceptual advantage.
This result aligns with prior perceptual studies showing that sarcasm recognition is facilitated when semantic content and intonation are congruent \citep{bryant2002recognizing, woodland2011context, bryant2010prosodic}. For instance, \citet{woodland2011context} found that incongruence between context and tone led to ambiguous or delayed judgments of sarcasm. 
Similarly, \citet{bryant2002recognizing, bryant2010prosodic} demonstrated that prosodic cues systematically signal sarcastic intent in spontaneous speech, particularly when linguistic cues alone are ambiguous.  

Taken together, these results highlight three key insights: (i) raw, general-purpose embeddings are insufficient and can reduce perceived expressivity; (ii) parameter-efficient adaptation with LoRA can inject semantic and pragmatic knowledge that enhances the recognition of sarcasm; and (iii)  retrieval-augmented conditioning further refines prosodic expressiveness by grounding synthesis in real sarcastic exemplars. In summary, these findings suggest the complementary roles of semantic and prosodic cues in sarcasm perception.


\section{Conclusion}


This work presents a unified computational framework for modeling sarcastic speech that captures how semantic and prosodic cues interact to convey sarcastic intent. Semantic representations are derived from a LoRA-adapted LLaMA 3, encoding discourse-level cues indicative of sarcasm, while prosodic exemplars retrieved via the RAG module provide patterns of intonation and emphasis characteristic of sarcastic delivery. By integrating these components, the framework enables the generation of speech that reflects the subtle interplay between meaning and prosody, producing utterances that are both intelligible and perceptually recognizable as sarcastic.

Beyond improvements in synthesis quality, our findings offer insights into the cognitive mechanisms underlying sarcasm perception. Results from both objective and subjective evaluations suggest that semantic and prosodic cues make complementary contributions to the recognition of sarcasm. 
This supports theoretical accounts of sarcasm as a pragmatic phenomenon arising from the interaction of meaning and delivery, rather than from isolated linguistic or acoustic features.

The proposed framework demonstrates how advances in LLMs and retrieval-based conditioning can be used to model subtle pragmatic phenomena. Unlike conventional emotional TTS systems that rely on coarse affective labels, our approach offers a more flexible and cognitively grounded way of controlling expressive speech through high-level semantic cues. Future work could further emphasize contextual information in sarcasm modeling, moving beyond isolated utterances toward discourse-aware detection and generation.

\printbibliography

\end{document}